\documentclass{article}

\usepackage{hyperref}
\usepackage{tikz}
\usepackage{amssymb}
\usepackage{wrapfig}
\usepackage{enumitem}
\usepackage{verbatim}

\title{Dataflow Matrix Machines and V-values: a Bridge between Programs and Neural Nets\thanks{For Festschrift in Honor of  L\'aszl\'o K\'alm\'an and Andr\'as Kornai.}}
\author{Michael Bukatin\thanks{HERE North America LLC, Burlington, MA, USA}
\and
Jon Anthony\thanks{Boston College, Chestnut Hill, MA, USA}}

\begin{document}

\maketitle

\begin{abstract}
{\em Dataflow matrix machines} generalize neural nets by replacing streams of numbers with streams of vectors (or other kinds
of {\em linear streams} admitting a notion of linear combination of several streams) and adding a few more changes on
top of that, namely arbitrary input and output arities for activation functions, countable-sized networks with finite dynamically
changeable active part capable of unbounded growth, and a very expressive {\em self-referential mechanism}.

While recurrent neural networks are Turing-complete, they form an esoteric programming platform,
not conductive for practical general-purpose programming. Dataflow matrix machines are more suitable as
a general-purpose programming platform, although it remains to be seen whether this platform can be
made fully competitive with more traditional programming platforms currently in use. At the same time,
dataflow matrix machines retain the key property of recurrent neural networks: {\em programs are
expressed via matrices of real numbers, and continuous changes to those matrices produce arbitrarily small variations in the programs
associated with those matrices}.

Spaces of vector-like elements are of particular importance in this context. In particular, we focus
on the vector space $V$ of finite linear combinations of strings, which can be also understood  as the vector space of
finite prefix trees with numerical leaves, the vector space of ``mixed rank tensors", or the vector space of
recurrent maps.

This space, and a family of spaces of vector-like
elements derived from it,
are sufficiently expressive to cover all cases of interest we are currently aware of, and allow
a compact and streamlined version of dataflow matrix machines based on a single
space of vector-like elements and variadic neurons. We call elements of these
spaces {\em V-values}. Their role in our context is somewhat similar to the role of S-expressions in Lisp.
\end{abstract}

\section{Introduction}

Andr\'as Kornai wrote his {\em Mathematical Linguistics} book~\cite{Kornai} while he and the first author of the present paper were working in the same office at MetaCarta
during the previous decade and sharing many fruitful moments. The book published 10 years ago was written as an introduction to the mathematical foundations of linguistics for computer
scientists, engineers, and mathematicians interested in natural language processing.

Dataflow matrix machines emerged 2 years ago~\cite{BukatinMatthewsDataflowGraphsAsMatrices} and a series of technical papers have been written on the subject since
then~\cite{BMR1, BMR2, BMR3, BMR4, BukatinAnthony}. The present paper is meant to be an introduction to the subject for researchers and engineers working in other fields.
We tried to keep the style of chapters and sections of {\em Mathematical Linguistics} in mind while writing this article.

\rule{320pt}{0.1pt}
\vspace{5pt}

Artificial neural networks are a powerful machine learning platform based on processing the streams of numbers.

It is long known that recurrent neural networks are expressive enough to encode any algorithm, if they are equipped with a reasonable form of unbounded 
memory~\cite{McCullochPitts, Pollack, SiegelmannSontag}.
There is a long history of synthesis of algorithms expressed as neural networks both by compilation and by machine learning methods.

However, conventional neural networks belong to the class of {\em esoteric programming languages} and do not constitute a convenient platform for
manual software engineering. 

In particular, there is a considerable history of using neural networks to synthesize and modify other neural networks, including self-modification.
However, the limitations of conventional neural networks as a software engineering platform make efforts of this kind quite challenging.

The main key point of the approach of {\em dataflow matrix machines} is that the natural degree of generality for the neural model of computations is not the streams of numbers,
but arbitrary {\em linear streams}. 

The other enhancements dataflow matrix machines make to the neural model of computations are neurons of {\em variable input and output arity}, 
novel models of unbounded memory based on countable-sized weight-connectivity matrices with finite number of non-zero weights at any given time and, more generally, on streams of
countable-dimensional vectors, and {\em explicit self-referential facilities}.

This results in a more powerful and expressive machine learning platform. 

When one considers  dataflow matrix machines as a software engineering framework,  it turns out that the restriction to linear streams
and to programs which admit continuous deformations
is less severe than one could have thought a priori given the discrete nature of conventional programming languages. 

Dataflow matrix machines are considerably closer to being
a general-purpose programming platform than recurrent neural nets, while retaining the key property of recurrent neural nets that large classes of programs can be parametrized by matrices of numbers, and therefore synthesizing appropriate matrices is sufficient to synthesize programs.

\paragraph{Linear streams.}

Dataflow matrix machines are built around the notion of linear streams. Generally speaking, we say that a space of streams is a {\em space of linear
streams}, if a meaningful notion of linear combination of several streams with numerical coefficients is well-defined.  

The simplest example of a space of linear streams is the space of sequences of numbers. A slightly more complicated example comes from
considering a vector space $V$ and the space of sequences of its elements, $(v_1, v_2, \ldots)$.

In the first few sections of this paper, this simple version of the notion of linear streams would be sufficient.
The discrete time is represented by non-negative integers, a particular vector space $V$ is fixed, and the space of functions from time to $V$
forms the space of linear streams in question.

To distinguish between streams based on different linear spaces, e.g. $V_1$ and $V_2$, we talk about different {\em kinds of
linear streams}.

In Section~\ref{sec:Linear}, we describe a sufficiently general notion of linear streams which includes, for example,
streams of samples from sequences of probability distributions over an arbitrary measurable space $X$.
All constructions in the present paper work in this degree of generality.

\paragraph{Neuron types.} The originally developed formalism of dataflow matrix machines was heavily typed~\cite{BMR2}. One considered a diverse collection of {\em kinds of
linear streams}, and a diverse collection of {\em types of neurons} with explicit fixed input and output arities.

The more recent version is close to being type free. It uses a single
kind of linear streams based on a ``sufficiently universal" space of
elements we call V-values. V-values enable the use of {\em variadic
  neurons} which have arbitrary input and output arities, eliminating
the need to keep track of input and output arities. The neuron types
still exist in that they have different activation functions.

\paragraph{Structure of this paper.} We start with an informal introduction to
recurrent neural networks and to the typed version of dataflow matrix machines (DMMs) in Section~\ref{sec:rnn-dmm}.
In Section~\ref{sec:V-values}, we present the theory of V-values.
In Section~\ref{sec:variadic}, we describe DMMs based on V-values and variadic neurons.
We discuss linear streams in Section~\ref{sec:Linear}. In that section we also discuss embeddings of discrete objects into
vector spaces, and into spaces of linear streams.

Programming patterns in DMMs are presented in Section~\ref{sec:programming}. A very expressive self-referential mechanism
which is a key element of our approach is presented in Section~\ref{sec:self-ref}.

The issues related to expressing the network topology are the subject of Section~\ref{sec:network-topology} and the issues
related to subnetworks and modularization are the subject of Section~\ref{sec:subnetworks}.

We discuss some of the potential approaches to using DMMs in machine learning in Section~\ref{sec:learning}. The concluding
Section~\ref{sec:Historical} contains historical remarks and discussion of related work.

\section{From Recurrent Neural Networks to Dataflow Matrix Machines}\label{sec:rnn-dmm}

The essence of artificial neural architectures is that linear and nonlinear transformations are interleaved.
Then one can control neural computations by only modifying the linear part and keeping the non-linear part fixed.

\begin{wrapfigure}{l}{0.5\textwidth}
\begin{tikzpicture}
   \clip (-2.0, -2.0) rectangle (4.0, 2.0);
   \draw [->] (-0.8, 0) -- (-0.8, 1) node [left] {$i_m$};
   \draw [->] (-1.6, 0) -- (-1.6, 1) node [left] {$i_1$};
   \filldraw (-1.2,0) circle [radius=0.5pt]
                 (-1.4,0)  circle [radius=0.5pt]
                 (-1.0, 0) circle [radius=0.5pt]
                 (-0.8, 0) circle [radius=1pt]
                 (-1.6, 0) circle [radius=1pt];

   \draw [->] (-0.4, -1) node [left] {$x_1$} -- (-0.4, 1) node [midway, above right] {$f_1$} node [right] {$y_1$};
   \draw [->] (0.4, -1)  node [left] {$x_k$} -- (0.4, 1) node [midway, above right] {$f_k$} node [right] {$y_k$};
   \filldraw (0,0) circle [radius=0.5pt]
                 (-0.2,0)  circle [radius=0.5pt]
                 (0.2, 0) circle [radius=0.5pt];

   \draw (0.8, -1) node [right] {$o_1$} -- (0.8, 0);
   \draw (1.6, -1) node [right] {$o_n$} -- (1.6, 0);
   \filldraw (1.2,0) circle [radius=0.5pt]
                 (1.4,0)  circle [radius=0.5pt]
                 (1.0, 0) circle [radius=0.5pt]
                 (0.8, 0) circle [radius=1pt]
                 (1.6, 0) circle [radius=1pt]; 

  \draw [->, very thick] (0, 1.2) .. controls (3.5, 4.2) and (3.5, -4.2) .. (0, -1.2)  node [midway, right] {{\bf W}};

\end{tikzpicture}
\caption{``Two-stroke engine" for an RNN. Figure from~\cite{BukatinAnthony}.}
\label{fig:rnn}
\end{wrapfigure}
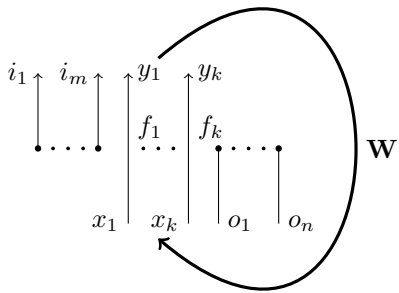

Therefore, neural architectures such as {\em recurrent neural networks} (RNNs) can be viewed as ``two-stroke engines" (Figure~\ref{fig:rnn}),
where the  ``two-stroke cycle" of a linear ``down movement" followed by a typically non-linear ``up movement" is repeated indefinitely. 

The network consists of the weight matrix {\bf W} and the neurons. The neuron $k$ has input and output streams of
numbers, $x^t_k$ and $y^t_k$, associated with it, where $t$ is discrete time. The network also has streams
of numbers $i^t_m$ representing external inputs, and streams of numbers $o^t_n$ representing external outputs.

On the ``down movement", neuron inputs and network external outputs are computed by applying linear transformation {\bf W} to the
neuron outputs and network external inputs: $(x_1^{t+1}, \ldots, x_k^{t+1}, o_1^{t+1}, \ldots, o_n^{t+1})^{\top}=$\newline
{\bf W}$ \cdot (y_1^{t}, \ldots, y_k^{t}, i_1^{t}, \ldots, i_m^{t})^{\top}$. On the ``up movement", the neurons calculate
their outputs from their inputs using activation functions $f_k$ which are built into each neuron $k$ and are usually
non-linear:  $y_1^{t+1} = f_1(x_1^{t+1}), \ldots, y_k^{t+1} = f_k(x_k^{t+1})$.

Note that the computations during the ``up movement" are local to the neuron in question, while the computations during the linear ``down movement"
are potentially quite global, as any neuron output might potentially be linked to any neuron input by a non-zero element of {\bf W}.

Now, moving from RNNs to {\em dataflow matrix machines} (DMMs), consider a finite or countable collection of {\em kinds of linear streams},
a finite or countable collection of {\em neuron types}, with every neuron type specifying non-negative integer number of inputs,
non-negative integer number of outputs, the kind of linear streams associated with each input and each output, and an
activation function transforming the inputs to the outputs.

Take a countable collection of neurons of each type, so that we have a countable-sized overall network. However, we'll make sure that only
a finite part of this network is active at any given time (similarly to only a finite part of the Turing machine tape having non-blank symbols at any given time), and that processing time and memory are only spent on working with the
currently active part, while the rest exists simply as potentially infinite address space.

The network consists of a countable-sized connectivity matrix {\bf W} and a countable-sized collection of neurons described in the
previous paragraph. The connectivity matrix {\bf W} depends\footnote{When {\bf W}
changes with time, this change can be controlled from the outside or by the network itself via the self-referential mechanism described in Section~\ref{sec:self-ref}.} on discrete time $t$.

The matrix element $w_{(i,C_k), (j,C_l)}^t$ is the weight linking the output $j$ of the neuron $C_l$ to the input $i$ of the neuron $C_k$ at the moment $t$.
We impose the condition that at any given moment of time $t$ only finite number of matrix elements $w_{(i,C_k), (j,C_l)}^t$
are non-zero. Hence the connectivity matrix is inherently sparse, and the structure of its non-zero weights determines
the actual connectivity pattern of the network at any given moment of time.

\begin{wrapfigure}{l}{0.7\textwidth}
\begin{tikzpicture}
   \clip (-3.5, -2.0) rectangle (7.0, 2.0);

  \filldraw (-3.2,0) circle [radius=0.5pt]
                (-3.0,0)  circle [radius=0.5pt]
                (-3.4, 0) circle [radius=0.5pt]; 

   \draw [->] (-1.2, 0) -- (-0.8, 1) node [right] {$y_{2,C_1}$};
   \draw [->] (-1.2, 0) -- (-1.6, 1) node [left] {$y_{1,C_1}$};
   \draw (-2.0, -1)  node [left] {$x_{1,C_1}$} -- (-1.2, 0);
   \draw (-1.2, -1)  node [below] {$x_{2,C_1}$} -- (-1.2, 0) node [left] {$f_{C_1}$};
   \draw (-0.4, -1)  node [right] {$x_{3,C_1}$} -- (-1.2, 0);

  \filldraw (0,0) circle [radius=0.5pt]
                (-0.2,0)  circle [radius=0.5pt]
                (0.2, 0) circle [radius=0.5pt];

   \draw (1.6, -1) node [left] {$x_{1,C_2}$} -- (2.0, 0)  node [left] {$f_{C_2}$};;
   \draw (2.4, -1) node [right] {$x_{2,C_2}$} -- (2.0, 0);
   \draw [->] (2.0, 0) -- (1.2, 1) node [left] {$y_{1,C_2}$};
   \draw [->] (2.0, 0) -- (2.0, 1) node [above] {$y_{2,C_2}$};
   \draw [->] (2.0, 0) -- (2.8, 1) node [right] {$y_{3,C_2}$};

  \filldraw (3.2,0) circle [radius=0.5pt]
                (3.0,0)  circle [radius=0.5pt]
                (3.4, 0) circle [radius=0.5pt];

  \draw [->, very thick] (0.5, 1.2) .. controls (5.5, 4.2) and (5.5, -4.2) .. (0.5, -1.2)  node [midway, right] {{\bf W}};

\end{tikzpicture}

\caption{``Two-stroke engine" for a standard DMM~\cite{BukatinAnthony}.
Two of its neurons, $C_1$ and $C_2$, are explicitly shown.} 
\label{fig:dmmold}
\end{wrapfigure}
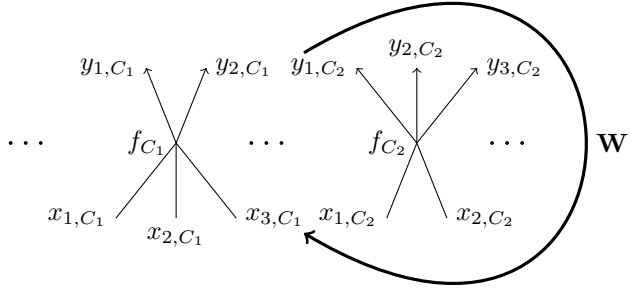

The DMM ``two-stroke engine" in a fashion similar to that of RNNs has a ``two-stroke cycle" consisting of a linear ``down movement" followed by
an ``up movement" performed by the activation functions of neurons (Figure~\ref{fig:dmmold}). This ``two-stroke cycle" is repeated indefinitely.  

``Down movement" is defined as follows.
For all inputs $x_{i,C_k}$ where there is a non-zero weight $w_{(i,C_k), (j,C_l)}^t$:

\begin{equation}\label{DMMlinearaction}
x_{i,C_k}^{t+1} = \sum_{\{(j,C_l) | w_{(i,C_k), (j,C_l)}^t \neq 0\}}w_{(i,C_k), (j,C_l)}^t * y_{j, C_l}^{t}.
\end{equation}

Note that
$x_{i,C_k}^{t+1}$ and $y_{j, C_l}^t$ may no longer be numbers, but elements of linear streams $x_{i,C_k}$ and $y_{j, C_l}$, so in order for Equation~\ref{DMMlinearaction} to be well-defined 
we impose the type correctness condition which states that
$w_{(i,C_k), (j,C_l)}^t$ is allowed to be
non-zero only if $x_{i,C_k}$ and $y_{j, C_l}$ belong to the same {\em kind} of linear streams\footnote{Recall that the number of inputs and outputs of a neuron $C$, the kind of
linear streams associated with each particular input or output of this neuron, and the built-in activation function $f_C$ of this neuron are determined by the type of the neuron in question.}.

We call a neuron $C$ active at the time $t$, if there is at least one non-zero connectivity weight from {\bf W} associated with one of its inputs or outputs.
Since {\bf W} has only a finite number of non-zero weights at any given time, there are only a finite number of active neurons in
the network at any given time.

``Up movement" is defined as follows. For all active neurons $C$: 
\begin{equation}
y^{t+1}_{1,C},...,y^{t+1}_{n,C} = f_C (x^{t+1}_{1,C},...,x^{t+1}_{m,C}).
\end{equation}

Here $m$ is the input arity of neuron $C$ and $n$ is its output arity, so $f_C$ has $m$ inputs and $n$ outputs.
 If $m=0$, then $f_C$ has no arguments.
If $n=0$, then the neuron just consumes data, and does not produce streams on the ``up movement".
Given that the input and output arities of neurons are allowed to be zero, special handling of network inputs and
outputs which was necessary for RNNs (Figure~\ref{fig:rnn}) is not needed here. The neurons responsible for
network input and output are included on par with all other neurons.

The resulting formalism is very powerful, and we discuss what can be done with it later in the paper. However, its
complexity is a bit unpleasant. The need to keep track of various kinds of linear streams and of the details of various neuron
types is rather tiresome. It would be great to have only one sufficiently expressive kind of linear streams, and, moreover,
to avoid the need to specify the arity of activation functions, while still enjoying the power of having multiple inputs and outputs
within a single neuron. The spaces of {\em V-values} discussed in the next section allow us to achieve just that, while further
increasing the power of a single neuron.

\section{V-values}\label{sec:V-values}

In this section, we define vector space $V$ which is sufficiently rich to represent vectors from many other spaces encountered in practice.

In Section~\ref{sec:extendedV}, we show how to enrich this construction in those situations where it is not sufficiently universal,
resulting in a family of vector spaces.

We call both the elements of these vector spaces and the hash-map-based representations of those elements {\em V-values}.
In our context, V-values play the role somewhat similar to the role
of S-expressions in the context of Lisp.

Implementation-wise, we create a version of S-expressions which is dictionary-based, rather than list-based. In this section, we require all
atoms of those dictionary-based S-expressions to be numbers.
A more general form of leaves in  V-values is considered in Section~\ref{sec:extendedV}.

Speaking more formally, we start with a finite or countable alphabet $L$ of labels (which we sometimes call tokens or keys). One can think about
elements of $L$ as words from some language defined over some other alphabet, which allows us to think about meaningful
languages of labels.

We are going to consider several equivalent ways to define tree-like structures with intermediate nodes labeled by
elements of $L$ and with leaves labeled by numbers. Some of these ways are ``depth-first", and they are easier
to present mathematically, and some are ``breadth-first", and they are more fundamental to us, as we use them
in our implementation and as they enable the use of variadic neurons. 

These tree-like structures
can be viewed as

\begin{itemize}[noitemsep]
  \item Finite linear combinations of finite strings;
  \item Finite prefix trees with numerical leaves;
  \item Sparse ``tensors of mixed rank" with finite number of non-zero elements;
  \item Recurrent maps from $V \cong \mathbb{R}\oplus (L \rightarrow V)$  admitting finite descriptions.
\end{itemize}

\subsection{Finite Linear Combinations of Finite Strings}\label{sec:finitestring-generated}

To start with ``depth-first" methods, consider space $L^*$ of finite sequences of elements of $L$ (including the empty
sequence), and construct $V$ as the vector space of formal finite linear combinations\footnote{the space of functions $f: L^* \rightarrow \mathbb{R}$
such that $f(w)\neq 0$ for no more than finite number of $w \in  L^*$; the operations are pointwise: $(f+g)(w) = f(w)+g(w)$ and $(\alpha f)(w) = \alpha f(w)$} 
of the elements of $L^*$ over $\mathbb{R}$.

We denote the empty sequence of elements of $L$ as $\varepsilon$, and we denote non-empty sequences of elements of $L$,
$(l_1, \ldots ,l_n)$, as $ l_1 \leadsto \ldots \leadsto l_n$. Since we are talking about formal finite linear combinations
of elements of $L^*$, we need a notation for the multiplication of real number $\alpha$ and generator $ l_1 \leadsto \ldots \leadsto l_n$,
$\alpha \cdot  (l_1 \leadsto \ldots \leadsto l_n)$.

For reasons, which become apparent in the next subsection, it is convenient to denote $\alpha \cdot  (l_1 \leadsto \ldots \leadsto l_n)$
as $l_1 \leadsto \ldots \leadsto l_n \leadsto \alpha$.

\subsection{Finite Prefix Trees with Numerical Leaves}\label{sec:prefix-trees}

One can think of $l_1 \leadsto \ldots \leadsto \l_n$ as a path in a prefix tree ({\em trie}), with intermediate nodes being labeled by letters from $L$.
So when one considers $\alpha \in \mathbb{R}$, one can express the presence of term $\alpha \cdot (l_1 \leadsto \ldots \leadsto l_n)$ in our
linear combination as presence of path with the intermediate nodes labeled by $l_1, \ldots, l_n$ and the leaf labeled
by $\alpha$. We denote this path as $l_1 \leadsto \ldots \leadsto l_n \leadsto \alpha$.
Because we have finite linear combinations (terms and paths corresponding to $\alpha = 0$ tend to be
omitted), we are talking about {\em finite prefix trees with numerical leaves}.

\paragraph{Example.} The linear combination 3.5 $\cdot$ ($\varepsilon$) + 2 $\cdot$ ({\tt :foo}) + 7 $\cdot$ ({\tt :foo :bar}) - 4 $\cdot$ ({\tt :baz :foo :bar}),  
i.e., ($\leadsto$ 3.5) + ({\tt :foo} $\leadsto$  2) + ({\tt :foo} $\leadsto$ {\tt :bar} $\leadsto$ 7) + ({\tt :baz} $\leadsto$ {\tt :foo} $\leadsto$ {\tt :bar} $\leadsto$ -4). 

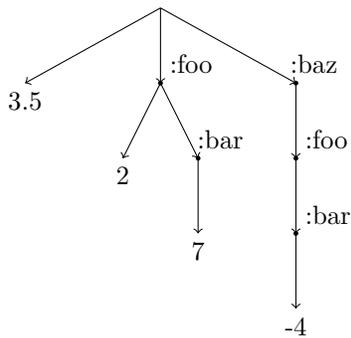
\begin{wrapfigure}{l}{0.6\textwidth}
\begin{tikzpicture}
  \clip (-3.0, -4.5) rectangle (3.0, 0.0);

  \draw[->](0, 0) -- (-1.8, -1)  node [below] {3.5};

  \draw[->](0, 0) -- (0, -1) node [midway,  below right] {:foo};

    \draw[->](0, -1) -- (0.5, -2) node[midway, below right] {\ :bar};

      \draw[->](0.5, -2) -- (0.5, -3) node [below] {7};

    \draw[->](0, -1) -- (-0.5, -2) node [below] {2};

  \draw[->](0, 0) -- (1.8, -1) node [midway, below  right] {\ \ \ \ \ \ :baz};

      \draw[->](1.8, -1) -- (1.8, -2) node [midway, below  right] {:foo};

      \draw[->](1.8, -2) -- (1.8, -3) node [midway, below  right] {:bar};      

      \draw[->](1.8, -3) -- (1.8, -4) node [below] {-4};

  \filldraw (1.8, -1) circle [radius=0.7pt]
                (1.8, -2) circle [radius=0.7pt]
                (1.8, -3) circle [radius=0.7pt]
                (0, -1) circle [radius=0.7pt]
                (0.5, -2) circle [radius=0.7pt];

\end{tikzpicture}

\caption{The prefix tree for this example.} 
\label{fig:trie}
\end{wrapfigure}

The empty string $\varepsilon$ with non-zero coefficient $\beta$, written as $\beta \cdot (\varepsilon)$ or simply $\leadsto \beta$, corresponds 
to the leaf with non-zero $\beta$ attached directly to the root of the tree.

The prefix tree (trie) for this example is shown in Figure~\ref{fig:trie}. In this particular example, we label intermediate nodes with
Clojure keywords, which start with ``{\tt :}" character. This example reminds us about the earlier remark that one can often think about
letters from alphabet $L$ as words from some language defined over some other, more conventional alphabet, and that this allows us to think about meaningful
languages of labels.

\subsection{Sparse ``Tensors of Mixed Rank" }

Yet another ``depth-first" way of looking at this situation is to consider $l_1 \leadsto \ldots \leadsto l_n \leadsto \alpha$ to be an element
of sparse multidimensional array with $n$ dimensions. E.g.  $l_1 \leadsto l_2 \leadsto \alpha$ is an element of a sparse matrix with its row
labeled by $l_1$ and its column labeled by $l_2$ and $\alpha$ being the value of the element. 

The  non-zero leaf attached directly to the root of the tree, 
$\beta \cdot (\varepsilon) = \leadsto \beta$, is considered to be a scalar with value $\beta$. 

Each string of length one with non-zero coefficient $\gamma$, that is $l \leadsto\gamma$ (the leaf with non-zero
$\gamma$ attached to the end of the path of length one with the intermediate node in the path labeled by $l$), is considered to be
a coordinate of a sparse array, where the coordinate is labeled by $l$ and has value $\gamma$. 

Each string (path) of length three with
non-zero coefficient $\gamma\,'$, written as $l_1 \leadsto l_2 \leadsto l_3 \leadsto \gamma\,'$, is considered to be an element of sparse three-dimensional
array, etc.

The standard convention
in machine learning is to call multidimensional arrays with $n$ dimensions ``tensors of rank $n$". Because our
linear combinations generally include sequences from $L^*$ of different lengths, we have to talk about {\em sparse
``tensors of mixed rank"}. 
For example, the vector ($\leadsto$ 3.5) + \linebreak ({\tt :foo} $\leadsto$  2) + ({\tt :foo} $\leadsto$ {\tt :bar} $\leadsto$ 7) + ({\tt :baz} $\leadsto$ {\tt :foo} $\leadsto$ {\tt :bar} $\leadsto$ -4)
from the previous subsection is the sum of scalar 3.5, sparse array with one non-zero element {\tt d1[:foo] = 2}, sparse 
matrix with one non-zero element {\tt d2[:foo, :bar] = 7}, and sparse three-dimensional array with one non-zero
element\linebreak {\tt d3[:baz, :foo, :bar] = -4}, so it is a typical ``tensor of mixed rank".

Therefore all usual vectors, matrices, and tensors of any dimension can be represented in $V$.
This is convenient from the viewpoint of machine learning as multidimensional tensors often occur in the machine
learning practice.

The space $V$ is a direct sum of the one-dimensional space of
scalars and spaces of n-dimensional arrays: $V = V_0 \oplus V_1
\oplus V_2 \ldots$ Since $L$ is countably infinite, $V_1, V_2, \ldots$ are
infinite-dimensional vector spaces. If $L$ is finite and consists of
$\mbox{Card}(L)$ elements, then the dimension of $V_i$ as a vector space is
$\mbox{Card}^i(L)$.

This is also a good point to transition to ``breadth-first" representations. A sparse matrix can also be viewed as a
map from elements of $L$ labeling its non-zero rows to the sparse vectors representing those rows (these sparse vectors
are maps from elements of $L$ labeling columns of the matrix to the values of the actual non-zero matrix elements;
zero elements are omitted, as usual).

\subsection{Recurrent Maps}\label{sec:recurrent}

To obtain a ``breadth-first" representation for a general element $v \in V$, we first note that there is a possibility
that the $\alpha \cdot (\varepsilon) = \leadsto \alpha$  belongs to $v$ with non-zero coefficient $\alpha$, in which case the corresponding
coordinate of $v$ is a non-zero ``tensor of rank 0", i.e. the scalar with value $\alpha$.

Then for each letter $\l_1 \in L$, such that some non-zero term $l_1 \leadsto l_2 \leadsto \ldots \leadsto l_n \leadsto \beta$ belongs to $v$, we consider all terms from $v$ which share the same first letter $l_1$, namely  $l_1 \leadsto l_2 \leadsto \ldots \leadsto l_n \leadsto \beta,\  l_1 \leadsto l'_2 \leadsto \ldots \leadsto l'_m \leadsto \gamma, \ldots$, and consider $v^{l_1} \in V$ consisting of those terms with the first letter removed, namely $l_2 \leadsto \ldots \leadsto l_n \leadsto \beta,\   l'_2 \leadsto \ldots \leadsto l'_m \leadsto \gamma, \ldots$

We map each such letter $l_1$ to $v^{l_1}$, and we map  each letter $l \in L$ for which $v$ does not have  a non-zero term starting from $l$ to zero vector (zero element of V).
The finite description of our map only needs to include the finite set of $\langle l_1, v^{l_1} \rangle$ pairs, and pairs $\langle l, 0 \rangle$ and $\langle l_1, 0 \rangle$ can all be omitted.

An element of $v$ is then a pair consisting of a scalar and a map from $L$ to $V$ admitting a finite description. Either or both elements of this pair can be zero.

As a vector space, $V$ satisfies the following equation:

\begin{equation}
V \cong \mathbb{R}\oplus (L \rightarrow V).
\end{equation}

Here $L \rightarrow V$ is a space of such maps from $L$ to $V$ that only a finite number of elements of $L$ map to non-zero elements of $V$.
Let's call such maps {\em finitary}. 
This equation is an {\em isomorphism of vector spaces}. It reflects the fact that every element $v \in V$ can be represented as a
pair of $\alpha \in \mathbb{R}$ and finitary map $l \mapsto v^l$, as was shown earlier in the present subsection, and vice versa
every pair consisting of $\alpha \in \mathbb{R}$ and a finitary map $L \rightarrow V$ is obtained in this fashion.

We would like to be able to represent elements of $V$ not by pairs of a number and a map, but simply by maps.
To include numbers $\alpha$ into the map itself, we would need a separate label for them that does not appear in $L$.
Thus we take a new key $n \not\in L$  and $L' = L \cup \{n\}$, then represent $\langle \alpha, \{ \langle l_1, v^{l_1} \rangle, \ldots, \langle l_n, v^{l_n} \rangle \} \rangle$ as $\{ \langle n, \alpha \rangle, \langle l_1, v^{l_1} \rangle, \ldots, \langle l_n, v^{l_n} \rangle \}$. 

Since $\mathbb{R}$ is embedded into $V$ via representation of $\alpha \in \mathbb{R}$ as $\langle \alpha, 0 \rangle$,
space $V$ is isomorphic to a subspace of $L' \rightarrow V$. This isomorphism is why we call this space a space of {\em recurrent maps}: every element of $V$ is represented as
a finitary map from extended alphabet $L'$ to the space $V$ itself,

This representation of $V$ via finitary maps to $V$ is fundamental to our constructions in the present paper,
because it translates directly to our Clojure implementation of core DMMs primitives~\cite{DMM} and
because it allows us to introduce {\em variadic neurons}.

\paragraph{Implementation.} We implement elements $v \in V$ as recurrent maps. Usually, programming languages provide
{\em dictionaries} or {\em hash-maps} suitable for this purpose. In our Clojure implementation, elements $v \in V$ are represented by
hash-maps, which map elements of $L'$ to $V$.

Typically, $L$ will be the set of all legal hash-map keys available in our language with the exception of
a few keys reserved for other purposes. In particular, we reserve Clojure keyword {\tt :number} to be mapped into the scalar component
of a pair $\langle \alpha, \{ \langle l_1, v^{l_1} \rangle, \ldots, \langle l_n, v^{l_n} \rangle \} \rangle$.
So, in our case $L' = L \cup \{${\tt :number}$\}$.

Therefore, $\langle \alpha, \{ \langle l_1, v^{l_1} \rangle, \ldots, \langle l_n, v^{l_n} \rangle \} \rangle$ is represented in Clojure by the hash-map
{\tt \{:number $\alpha,  l_1\,\, v^{l_1}, \ldots, l_n\,\, v^{l_n}$\}}.

Here $v^{l_1}, \ldots, v^{l_n}$ are represented by similar hash-maps themselves resulting in {\em nested hash-maps}, and keys from $L$ can have
rather complex structure, if desired, taking advantage of great variety of hash-map keys allowed in Clojure. 

When element $v \in V$ is simply a scalar (the pair $\langle \alpha, 0 \rangle$), the implementation is allowed to simply use number
$\alpha$ instead of the hash-map {\tt \{:number $\alpha$\}}.

\paragraph{Example from Section~\ref{sec:prefix-trees}.} The sum ($\leadsto$3.5) + ({\tt :foo} $\leadsto$  2) +
({\tt :foo} $\leadsto$
{\tt :bar} $\leadsto$ 7) + ({\tt :baz} $\leadsto$ {\tt :foo} $\leadsto$ {\tt :bar} $\leadsto$ -4) is represented as Clojure hash-map
{\tt \{:number 3.5, :foo \{:number 2, :bar 7\}, :baz \{:foo \{:bar -4\}\}\}}.

\paragraph{Variadic Neurons.} We use the formalism of V-values to eliminate the need to keep track of the number of
input and output arguments of the activation functions. We describe variadic neurons and DMMs based on
variadic neurons in Section~\ref{sec:variadic}.

To conclude the present section, V-values are essentially a dictionary-based version of S-expressions.
Section~\ref{sec:extendedV} removes the restriction that all atoms must be numbers and allows to incorporate
complex objects under reserved keywords.

\section{Variadic Neurons}\label{sec:variadic}

The activation functions of variadic neurons transform a single
stream of V-values into a single stream of V-values.

However, the labels at the first level of those V-values are dedicated to serve as the
names of input and output arguments. Therefore, a neuron is a priori {\em variadic} and can potentially handle
a countable collection of inputs and produce a countable collection of outputs (although our usual restrictions
of keeping the active part of the network finite would in practice limit those collections to finite at any given moment
of time).

Here is an example of an activation function for neuron with two arguments, $x$ and $y$, outputting two
results, difference ($x-y$)  and negative difference ($y-x$). 
This and all subsequent code examples in this paper are written in Clojure~\cite{Clojure}.
Assume for the purpose of this example that {\tt my-minus} function is available to compute these 
subtractions of one V-value from another V-value\footnote{Generally, one would want to define {\tt my-minus} in terms of DMM core primitives for V-values implemented in~\cite{DMM}:
{\tt(defn my-minus [x y] (rec-map-sum x (rec-map-mult -1 y))) }}:

\begin{verbatim}
    (defn symmetric-minus [input]
      (let [x (get input :x {})
            y (get input :y {})]
        {:difference (my-minus x y)
         :negative-difference (my-minus y x)}))
\end{verbatim}

The {\tt input} is a hash-map, representing a V-value. The arguments are subtrees of {\tt input} corresponding to {\tt :x,:y} $\in L$. 
These subtrees are computed by the
expressions {\tt (get input :x \{\})} and {\tt (get input :y \{\})}.
If the subtree in question is not present, the empty hash-map {\tt \{\}} is returned.
The empty hash-map represents zero vector (zero element of $V$) in our implementation.
The function outputs a V-value with two subtrees corresponding to {\tt :difference,:negative-difference} $\in L$.

So, the arguments are combined into a single V-value, {\tt input}, and the outputs are combined into a single
return V-value.

The non-trivial aspect of this approach is that instead of mapping neuron outputs to neuron inputs, the network matrix {\bf W} now
maps subtrees at the first level of the neuron outputs to subtrees at the first level of the neuron inputs. 
We shall see in the present section that the network matrix {\bf W} naturally acquires a structure
of multidimensional tensor under this approach.

\subsection{Space $U$}

Because we are going to use the keys at the first level of V-values as names of inputs and outputs, we don't want any top-level leaves, that
is we don't want non-zero scalars (``tensors of rank 0") in our V-values. Hence we'll be using space $U = L \rightarrow V$, namely we'll use values
$u \in U$ as {\em inputs and outputs of the neuron activation function} (which will always have arity one), and those values $u$ will contain {\em actual inputs and outputs of the neuron}
at their first level.

Observe that $V \cong \mathbb{R}\oplus (L \rightarrow V) = \mathbb{R}\oplus U$, and that $U$ therefore satisfies the equation
$U \cong L \rightarrow (\mathbb{R}\oplus U)$.

The  activation functions of the neurons map $U$ to $U$,
transforming single streams of elements of $U$. The labels at the first level
of the elements of $U$ serve as names of inputs and outputs.

\subsection{Multidimensional Structure on $W$}

The network matrix {\bf W} must provide a linear map from the concatenation of the first levels of
elements of $U$ which are the outputs of all neurons, to the concatenation of the first level of
elements of $U$ which are the inputs of all neurons. In our example above, a neuron with {\tt symmetric-minus}
activation function would have two V-values at its input, one labeled by {\tt :x} and one labeled by {\tt :y}, assembled into
one input map.
Such a neuron would also have two V-values at its output, one labeled by {\tt :difference} and
one labeled by {\tt :negative-difference}, assembled into one output map.

However, we consider an infinite collection of V-values on input of each neuron, with V-values
labeled by all elements of $L$, and an infinite collection of V-values on output of each neuron, with V-values
labeled by all elements of $L$, since nothing restricts activation functions from using
any of those labels.

Now we want to take infinite collections of V-values on {\em output} of each neuron for all neurons and
join those infinite collections together into a single infinite collection of V-values, and we shall apply matrix {\bf W}
to this unified infinite collection (imposing the usual condition about only a finite number of
relevant elements or vectors being non-zero at any given moment of time).

We also want to take infinite collections of V-values on {\em input} of each neuron for all neurons and
join those infinite collections together into a single infinite collection of V-values, and matrix {\bf W}
will produce this unified infinite collection each time it is applied to the collection described in the previous
paragraph.

Below we follow
closely the material from Section 3.2 of~\cite{BukatinAnthony}.

Consider {\em one input} situated on the first level of the element of $U$ serving as the
argument of the activation function for one neuron, and the row of the network matrix {\bf W}
responsible for computing {\em that input} from the concatenation of the first levels of
elements of $U$ which are the outputs of all neurons.

The natural structure of indices of this row is not flat, but hierarchical. At the very least,
there are two levels of indices: neurons and their outputs.

We currently use three levels of hierarchy in our implementation: neuron types (which are Clojure {\em vars} referring to implementations of
activation functions $U \rightarrow U$), neuron names, and names of the outputs.  Hence, matrix rows are three-dimensional sparse arrays (sparse ``tensors of rank 3")  in our current implementation. 

The natural structure of indices of the array of rows is also not flat, but hierarchical. At the very least,
there are two levels of indices: neurons and their inputs.
We currently use three levels of hierarchy in our implementation: neuron types, neuron names, and
names of the inputs.

Hence, the network matrix {\bf W} is a six-dimensional
sparse array (sparse ``tensor of rank 6")  in our current implementation.

\subsection{DMMs Based on V-values and Variadic Neurons}

\begin{wrapfigure}{l}{0.7\textwidth}
\begin{tikzpicture}
   \clip (-3.5, -2.0) rectangle (7.0, 2.0);

  \filldraw (-3.2,0) circle [radius=0.5pt]
                (-3.0,0)  circle [radius=0.5pt]
                (-3.4, 0) circle [radius=0.5pt]; 

  \draw [->] (-2.6, -1.5) node[right] {$x_{f, n_f}$} -- (-2.6, 1.5) node [midway, above right] {$f$} node[right] {$y_{f, n_f}$};

  \filldraw (0,0) circle [radius=0.5pt]
                (-0.2,0)  circle [radius=0.5pt]
                (0.2, 0) circle [radius=0.5pt];

  \draw [->] (0.6, -1.5) node[right] {$x_{g, n_g}$} -- (0.6, 1.5) node [midway, above right] {$g$} node[right] {$y_{g, n_g}$};

  \filldraw (3.2,0) circle [radius=0.5pt]
                (3.0,0)  circle [radius=0.5pt]
                (3.4, 0) circle [radius=0.5pt];

  \draw [->, very thick] (1.1, 1.1) .. controls (5.5, 4.3) and (5.5, -4.0) .. (1.1, -0.8)  node [midway, right] {{\bf W}};

  \foreach \y in {-1.0, 1.0}
    {

      \draw [densely dotted] (0.45, \y+0.15) ellipse [x radius=100pt, y radius=4pt];

     \foreach \x in {-1.0, 2.2}
       {

        \foreach \d in {-0.4, -0.15, 0.1}
           {
               \draw(\x-0.15, \y + 0.45) -- (\x+\d, \y+0.15); 
               \draw (\x+\d, \y + 0.15) -- (\x+\d-0.08, \y-0.15);
               \draw (\x+\d, \y + 0.15) -- (\x+\d+0.08, \y-0.15);
               \filldraw (\x+0.5*\d-0.05, \y-0.25) circle [radius=0.2pt];
               \filldraw (\x+0.5*\d+0.5, \y+0.15) circle [radius=0.2pt];
           }
       } 
     }
 
\end{tikzpicture}

\caption{``Two-stroke engine" for a DMM based on variadic neurons. Two neurons, $n_f$ and $n_g$, are explicitly pictured.
Their inputs and outputs, $x_{f, n_f}, x_{g, n_g}, y_{f, n_f},y_{g, n_g}$, are depicted as trees belonging to $U$.
{\bf W} is a linear map from the concatenation of the first levels of all $y_{h, n_h}$ trees to the concatenation of the first levels of all $x_{h, n_h}$ trees (Equation~\ref{eq:main}).} 
\label{fig:dmmnew}
\end{wrapfigure}
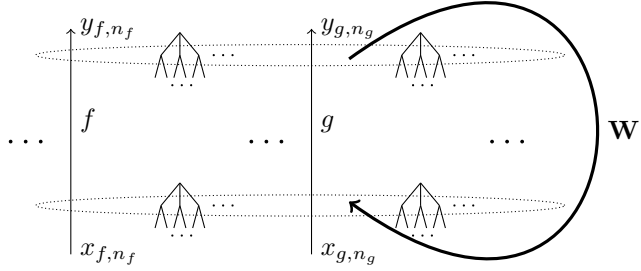

The DMM is a ``two-stroke engine" similar to that of
Section~\ref{sec:rnn-dmm}, and consists of a linear ``down movement"
followed by an ``up movement" performed by the activation functions of
neurons. This ``two-stroke cycle" is repeated indefinitely (Figure~\ref{fig:dmmnew}).

We allow names for neurons and their inputs and outputs to be any elements of $L$.
The address space is such that the network is countably-sized, but since the network matrix {\bf W}
has only a finite number of non-zero elements at any given time, and elements
of $U$ have only a finite number of non-zero coordinates at any given
time, we are always working with finite representations.

The network matrix {\bf W} (sparse ``tensor of rank 6") depends on $t$, and its element
$w_{f, n_f, i;\, g, n_g, o}^t$ is non-zero, if the output $o$ of neuron $n_g$ with the built-in activation function $g$
is connected to the input $i$ of neuron $n_f$ with the built-in activation function $f$ at the moment of time $t$, with number 
$w_{f, n_f, i;\, g, n_g, o}^t$ being the non-zero weight of this connection.

On the ``down movement", the network matrix $(w_{f, n_f, i;\, g, n_g, o}^t)$
is applied to an element of $U$ which contains all outputs of all neurons. The result
is an element of $U$ which contains all inputs of all neurons to be used during
the next ``up movement".
Each of those inputs is computed using the following formula:

\begin{equation}\label{eq:main}
x_{f, n_f, i}^{t+1} = \sum_{g \in F} \sum_{n_g \in L} \sum_{o \in L} w_{f, n_f, i;\, g, n_g, o}^t * y_{g, n_g, o}^t.
\end{equation}

Indices $f$ and $g$ belong to the set of neurons types $F$, which is simply the set of
transformations of $U$. Potentially, one can implement countable number of
such transformations in a given programming language, but at any given time only finite number
of them are defined and used. Indices $n_f$ and $n_g$ are the names of input and output neurons,
and indices $i$ and $o$ are the names of the respective input and output arguments of those neurons.

In the formula above, $w_{f, n_f, i;\, g, n_g, o}^t$ is a number (the connection weight), and $x_{f, n_f, i}^{t+1}$ and $y_{g, n_g, o}^t$ are elements of $V$ (not necessarily of $U$,
since the presence of scalars is allowed at this level).
The product $w_{f, n_f, i;\, g, n_g, o}^t * y_{g, n_g, o}^t$ multiplies vector $y_{g, n_g, o}^t$ by real number $w_{f, n_f, i;\, g, n_g, o}^t$.

This operation is performed for all  $f \in F$, all $n_f \in L$, all input names $i \in L$, such that
the corresponding three-dimensional row of {\bf W} has some non-zero elements at time $t$.

The result is finitely sized map $\{f \mapsto \{n_f \mapsto x_{f, n_f}^{t+1}\}\}$, where each $x_{f, n_f}^{t+1}$
is a finitely sized map from the names of neuron inputs to the values of those inputs, $\{i \mapsto x_{f, n_f, i}^{t+1}\}$.

On the ``up movement", each $f$ is simply applied to the elements of $U$ representing the single inputs
of the activation function $f$. This application is performed for all neurons $\langle f, n_f \rangle$ which are
present in the finitely sized map described in the previous paragraph\footnote{The neuron is fully determined by a pair $\langle f, n_f \rangle$. For each $f \in F$, there can be many
active neurons $\langle f, n_f \rangle$ with different $n_f$. For each $n_f \in L$, there can be many
active neurons $\langle f, n_f \rangle$ with different $f$.}:

\begin{equation}
y_{f, n_f}^{t+1} = f(x_{f, n_f}^{t+1}).
\end{equation}

An example of an activation function of a variadic neuron was given in the beginning of the present section.
For more examples of this kind see Section~\ref{sec:lin-mult} and Section~\ref{sec:datastruct}.

We gave the construction for the space $V$, but similar constructions also work for more general variants described in Section~\ref{sec:extendedV}.

\section{Linear Streams}\label{sec:Linear}

Informally speaking, we say that a space of streams is a {\em space of linear
streams}, if a meaningful notion of linear combination of several streams with numerical coefficients is well-defined.  

We would like to formalize this notion, while keeping the following examples in mind:

\begin{itemize}[noitemsep]
\item The space of sequences of numbers;

\item For a vector space $V$, the space of sequences of its elements, $(v_1, v_2, \ldots)$;

\item For a measurable space $X$, spaces of samples and signed samples drawn from $X$.
\end{itemize} 

We consider linear streams in somewhat limited generality here. First of all, we consider discrete sequential time, while other models of time, e.g. continuous, are also potentially of interest\footnote{We understand streams as functions from time to a set of objects, so that an object corresponds to
any given moment of time. In our paper, time tends to have a starting point, to be discrete, and to continue indefinitely, so we usually
model time by non-negative integers starting from 0 or 1.}. We consider linear streams over real numbers, while other systems of coefficients (especially, complex numbers) can also be quite fruitful.
Finally, we ground each space of linear streams in a vector space, while one could consider a more abstract approach to the notion of linear combination, where one works solely with
streams of abstract representations without grounding them in a vector space.

To define a particular {\em kind} of linear streams $k$ in a more formal manner, we specify background vector space $V_k$ and streams of
{\em approximate representations} of the underlying vectors from $V_k$. 
The approximate representations provide some information about the underlying vectors. Moreover, for every kind of linear streams $k$, we specify
a procedure computing an approximate representation of a linear combination $\alpha_1 v_{1,k} + \ldots + \alpha_n v_{n,k}$ from
approximate representations of vectors $v_{1,k}, \ldots, v_{n,k}$. We say informally that the approximate representations in question belong to a vector-like space
and we call them {\em vector-like elements}.

\subsection{Streams of Samples and Signed Samples}\label{sec:samples}

First, let's consider streams of samples from sequences of probability distributions over an arbitrary measurable space $X$ and their
linear combinations with positive coefficients.

 A sequence of probability distributions, $(\mu_1, \mu_2, \ldots)$,
can be represented by a sequence of elements of $X$ sampled from those distributions, $(x_1, x_2, \ldots)$. Note that $X$
is not required to be a vector space. Consider
$0 < \alpha < 1$ and sequences of probability distributions $(\mu_1, \mu_2, \ldots)$ and $(\nu_1, \nu_2, \ldots)$
represented by streams of samples $(x_1, x_2, \ldots)$ and $(y_1, y_2, \ldots)$. Produce a stream of samples
representing the sequence of probability distributions $(\alpha*\mu_1 + (1-\alpha)*\nu_1, \alpha*\mu_2 + (1-\alpha)*\nu_2, \ldots)$
as follows. Sample a random number uniformly from [0,1], and if this number is smaller than $\alpha$, pick $x_1$ as the first element
of our stream, otherwise pick $y_1$. Repeat this procedure for $x_2$ and $y_2$, and so on. Let's call this procedure a {\em stochastic
linear combination} of the streams of samples.

To include this example into our framework, we need a background vector space, and we need stochastic linear combinations
with positive and negative coefficients to be well-defined.

Let's observe that probability distributions over some measurable space $X$ belong
to the vector space of finite signed measures\footnote{measures which can take any finite real values, including negative values} over $X$.
Let's consider {\em signed samples}, that is, samples marked as being positive or negative.

Streams of signed samples are mentioned in Section 1.2 of~\cite{BM2015}. Here we follow
a more detailed treatment of them given in Appendix A.1 of~\cite{BukatinAnthony}. The underlying vector space is the space of all finite signed measures
over arbitrary measurable space $X$, and samples are pairs $\langle x, s \rangle$, with $x \in X$ and flag $s$ taking values -1 and 1. 

One considers streams of finite signed measures over $X$,  $\mu_1, \ldots, \mu_n$,
and streams of corresponding samples, $\langle x_1, s_1 \rangle, \ldots, \langle x_n, s_n \rangle$.

The procedure of computing a sample representing a signed measure
$\alpha_1 * \mu_1 + \ldots + \alpha_n * \mu_n$ is as follows. We pick index $i$ with
probability $\mid\alpha_i\mid / \sum_j \mid\alpha_j\mid$ using the absolute values of coefficients $\alpha$, and we pick
the sample $\langle x_i, \mbox{sign}(\alpha_i) * s_i \rangle$ (reversing the flag if the selected value $\alpha_i$ is negative) to represent
the linear combination $\alpha_1 * \mu_1 + \ldots + \alpha_n * \mu_n$.

For further discussion of expressive power of signed measures and signed samples see Section 1.2 of~\cite{BM2015}.
Issues related to missing samples and zero measures are mentioned in Appendix A.2 of~\cite{BukatinAnthony}.
A generalization to complex-valued measures and linear combinations with complex coefficients is considered in
the design notes for~\cite{DMM}\footnote{\url{https://github.com/jsa-aerial/DMM/blob/master/design-notes/Early-2017/sampling-formalism.md}}.

\subsection{Embedding}\label{sec:embedding}

The ability to represent characters, words, and other objects of discrete nature as vectors is one of the cornerstones
of success of modern neural networks.

The embedding of characters into a vector space generated by the alphabet (``one-hot encoding") is a basis for
rather spectacular results obtained by modern forms of recurrent neural networks, such as LSTMs~\cite{Karpathy}. The ability to learn
an optimal embedding of words into vectors is an integral part of a number of applications of recurrent neural nets to linguistics~\cite{Mikolov}.

When it comes to representing compound structures in vector spaces, there is obviously a lot of freedom and variety.
Throughout this paper, we work with embeddings of classes of dataflow matrix machines\footnote{considering each time a class of DMMs over some signature of neuron types} into corresponding vector spaces. One can argue that a class of
dataflow matrix machines forms a sufficiently rich space of objects, and that since one finds a meaningful natural embedding
of such a space into a vector space, one should expect to be able to find meaningful natural embeddings for a large variety
of spaces of objects.

As the previous subsection indicates, the notion of embedding into linear streams is more general
than the notion of embedding into vector spaces. As long as one is willing to consider a stream of objects of arbitrary
nature as drawn from some sequence of probability distributions over those objects, this stream belongs to a space
of linear streams of samples equipped with the stochastic version of  linear combination.

Hence one is able to obtain an embedding of a stream of objects into a space of linear streams without embedding individual objects into a vector space.

\subsection{A Family of Spaces of V-values}\label{sec:extendedV}

The space $V$ is very expressive, but sometimes it is not enough. If one wants to accommodate vectors from
some other vector space $V'$, a convenient way to do so is to allow elements of $V'$ in the leaves of the prefix trees. Usually one
still wants to be able to have just numbers inside the leaves as well, so one uses $\mathbb{V} = \mathbb{R}\oplus V'$ for
the space of leaves.

This means that elements of $\mathbb{V}$ are now used as coefficients $\alpha$ instead of real numbers in 
$\alpha \cdot (l_1 \ldots l_n) = l_1 \leadsto \ldots \leadsto l_n \leadsto \alpha$.
Considering a basis in $\mathbb{V}$, one can see that this whole construction corresponds to tensor product
$V \otimes \mathbb{V}$.

Elements of $\mathbb{V}$ are also used as elements of sparse ``tensors of mixed rank". The equation $V \cong \mathbb{R}\oplus (L \rightarrow V)$
becomes $V \cong \mathbb{V} \oplus (L \rightarrow V)$.

Implementation-wise, if, for example, $\mathbb{V}= \mathbb{R}\oplus V' \oplus V'' \oplus V'''$, then in addition to
the reserved key {\tt :number}, one would reserve additional keys for each of the additional components $V', V'', V'''$
to incorporate the non-zero instances of those components into leaves (and into hash-maps).

For example, Appendix A.3 of~\cite{BukatinAnthony} shows how to accommodate signed samples discussed in Section~\ref{sec:samples} above
within this framework. One considers
$\mathbb{V} =\mathbb{R}\oplus M$,
where $M$ is the space of finite signed measures over $X$, and one uses the reserved keyword {\tt :sample}
to incorporate signed samples into the leaves as necessary.

So, in this fashion a space $V \cong \mathbb{V} \oplus (L \rightarrow V)$ will still be represented by nested hash-maps,
and we are still going to call those nested hash-maps {\em V-values}.

\section{Programming}\label{sec:programming}

Here we discuss some of the programming patterns in dataflow matrix machines. Further programming tools
come from the presence of self-referential mechanism (Section~\ref{sec:self-ref}) and from modularization
facilities (Section~\ref{sec:subnetworks}).

\subsection{Linear and Multiplicative Constructions}\label{sec:lin-mult}

Linear and multiplicative constructions in dataflow matrix machines are well-covered in~\cite{BMR3}. The most
fundamental of them is a neuron with {\em identity activation function}. Consider some argument name, e.g. {\tt :accum}.
If we connect output {\tt :accum} of such a neuron
to its input {\tt :accum} with weight 1, this neuron becomes an {\em accumulator}. It adds together and accumulates in its
{\tt :accum} arguments the contributions
to its input {\tt :accum} made during each ``two-stroke cycle" by all other outputs in the network connected to the input {\tt :accum} of our neuron by non-zero weights.

Among multiplicative constructions, the most fundamental one is multiplication of an otherwise computed
neuron output by the value of one of its scalar inputs. This is essentially a fuzzy conditional, which can selectively
turn parts of the network on and off in real time via multiplication by zero, attenuate or amplify the signal, reverse the signal
via multiplication by -1, redirect
flow of signals in the network, etc. For further details see~\cite{BMR3}.

\begin{wrapfigure}{l}{0.58\textwidth}
\begin{tikzpicture}
   \clip (-2.0, -2.0) rectangle (7.0, 2.0);

  \filldraw (0,0) circle [radius=0.5pt]
                (-0.2,0)  circle [radius=0.5pt]
                (0.2, 0) circle [radius=0.5pt];

  \draw [->] (0.6, -1.5) node[right] {$x_{{\tt accum}, {\tt :my\mbox{\footnotesize -}neuron}}$} -- 
                    (0.6, 1.5) node [midway, above right] {{\tt accum}} node[right] {$y_{{\tt accum}, {\tt :my\mbox{\footnotesize -}neuron}}$};

  \filldraw (3.2,0) circle [radius=0.5pt]
                (3.0,0)  circle [radius=0.5pt]
                (3.4, 0) circle [radius=0.5pt];

  \draw [->] (1.9, 1.1) .. controls (5.5, 4.3) and (5.5, -4.0) .. (2.15, -0.85)  node [midway, left] {{1.0}};

  \draw [->] (1.7, -0.2) -- (1.5, -0.9);
  \filldraw (1.5, -0.25) circle [radius=0.2pt];
  \filldraw (1.6, -0.25) circle [radius=0.2pt];
  \filldraw (1.4, -0.25) circle [radius=0.2pt];
  \draw [->] (1.3, -0.2) -- (1.5, -0.8);

  \foreach \y in {-1.0}
    {

      \draw [densely dotted] (1.0, \y+0.15) ellipse [x radius=80pt, y radius=5pt];

     \foreach \x in {2.0}
       {

        \foreach \d in {-0.4}
           {
               \draw(\x-0.15, \y + 0.45) -- (\x+\d, \y+0.15) node [left] {\scriptsize{\tt :delta}}; 
               \draw (\x+\d, \y + 0.15) -- (\x+\d-0.08, \y-0.15);
               \draw (\x+\d, \y + 0.15) -- (\x+\d+0.08, \y-0.15);
               \filldraw (\x+0.5*\d-0.05, \y-0.25) circle [radius=0.2pt];
           }
       \foreach \d in {0.1}
           {
               \draw(\x-0.15, \y + 0.45) -- (\x+\d, \y+0.15) node [right] {\scriptsize{\tt :accum}}; 
               \draw (\x+\d, \y + 0.15) -- (\x+\d-0.08, \y-0.15);
               \draw (\x+\d, \y + 0.15) -- (\x+\d+0.08, \y-0.15);
               \filldraw (\x+0.5*\d-0.05, \y-0.25) circle [radius=0.2pt];
           }
        \foreach \d in {-0.4, -0.15, 0.1}
           {
              \filldraw (\x+0.5*\d-0.05, \y-0.25) circle [radius=0.2pt];
           }
       } 
     }

  \foreach \y in {0.9}
    {

      \draw [densely dotted] (1.0, \y+0.15) ellipse [x radius=80pt, y radius=5pt];

     \foreach \x in {2.0}
       {

        \foreach \d in {-0.15}
           {
               \draw(\x-0.15, \y + 0.45) -- (\x+\d, \y+0.15) node [right] {\scriptsize{\tt :single}}; 
               \draw (\x+\d, \y + 0.15) -- (\x+\d-0.08, \y-0.15);
               \draw (\x+\d, \y + 0.15) -- (\x+\d+0.08, \y-0.15);
 
           }
        \foreach \d in {-0.4, -0.15, 0.1}
           {
              \filldraw (\x+0.5*\d-0.05, \y-0.25) circle [radius=0.2pt];
           }
       } 
     }
 
\end{tikzpicture}

\caption{Connectivity of a neuron \hspace{80pt} {\tt [accum :my-neuron]}  with activation function
{\tt accum}, neuron name {\tt :my-neuron}, and input arguments
{\tt :accum} and {\tt :delta}, when this neuron is used as an accumulator.} 
\label{fig:accum}
\end{wrapfigure}
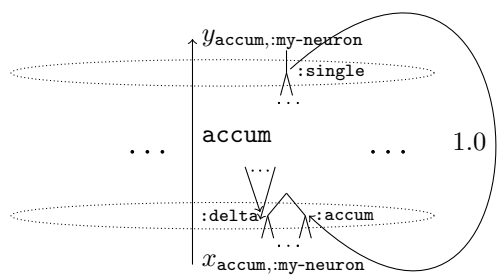

These facilities are quite powerful even for scalar flows of reals, and even more so for vector flows.
The lack of these facilities hinders the approaches which insist on only having non-linear activation functions,
or on only having neurons with a single input.

 Sometimes, it is convenient to take advantage of having
multiple inputs and use a neuron with $+$ activation function, $y = x + \Delta x$, as an accumulator, 
connecting $y$ to $x$ with weight 1 and accepting contributions from other outputs in the network
on input $\Delta x$~\cite{BMR4}. For details of this connectivity pattern in the current DMM architecture with
variadic neurons see Figure~\ref{fig:accum} .

The code for the activation function {\tt accum} of this accumulator looks as follows~\cite{DMM}:
\begin{verbatim}
    (defn accum [input]
      {:single (rec-map-sum (input :accum) (input :delta))})
\end{verbatim}

The function {\tt accum} takes a V-value {\tt input} as  argument, then it adds together the values of the two neuron inputs\footnote{{\tt rec-map-sum} is
one of DMM core primitives for V-values implemented in~\cite{DMM} which performs addition of V-values},
{\tt (input :accum)} and {\tt (input :delta)},
 obtaining the desired result, and then returns {\tt \{:single result\}} as the output. It can add together arbitrary elements of $V$.

In this example, the {\tt :accum} name for a neuron input stands for $x$, the {\tt :delta} name for a neuron
input stands for $\Delta x$, and the {\tt :single} is the name
of the only output this particular neuron has.

If we want a particular neuron {\tt :my-neuron} with built-in activation function {\tt accum} to work as an accumulator,
then the element of the network matrix connecting the output {\tt [accum :my-neuron :single]} together with the input
{\tt [accum :my-neuron :accum]} should be 1 (see Figure~\ref{fig:accum}). The corresponding matrix element
$w_{\tt accum, :my\mbox{\footnotesize -}neuron, :accum;\ accum, :my\mbox{\footnotesize -}neuron, :single}$ is expressed as

\begin{verbatim}
  {v-accum {:my-neuron {:accum
    {v-accum {:my-neuron {:single 1}}}}}}
\end{verbatim}

\noindent
in our current Clojure implementation, where {\tt v-accum} is defined as

\begin{verbatim}
  (def v-accum (var accum))
\end{verbatim}

\noindent
because the value of {\tt accum} itself used as a hash-key is not stable from recompilation to recompilation,
whereas {\tt (var accum)} is stable.

The contributions from other neurons are accepted on the input {\tt :delta}.

\subsection{Sparse Vectors}

The ability to handle sparse vectors in a straightforward manner is not available in scalar-based neural networks.
One has to allocate a neuron for every coordinate, and there is no straightforward way to avoid processing
those coordinates which happen to be zero at the moment.

In DMMs one can handle compact sparse vectors of very high, or even infinite dimensions, and the network
size does not depend on those dimensions. An example of a compact DMM built around an accumulator of sparse vectors
in the vector space generated by a given alphabet in order to keep track of the number of occurrences of each character
in a given string is studied in detail in~\cite{BMR3}. The savings can be drastic, as, for example, Unicode alphabet exceeds
100,000 characters, and with sparse representation only the actually occurring characters are stored and processed.

The same technique can be applied to keeping track of the number of occurrences of each word
in the text, where the underlying vector space generated by all possible words is infinitely-dimensional.

\subsection{Data Structures}\label{sec:datastruct}

An earlier paper~\cite{BMR3} focuses on allocating data structures in the body of the network itself, an approach
encouraged by the network's potentially infinite size and powerful self-modification facilities.

However, with the ability to process complicated vectors such as V-values, it is natural to encode and process
data structures on that level.
The use of structure sharing immutable data structures, the
default in Clojure, should make passing complicated structures
through ``up movements" and ``down movements" reasonably
efficient, as seen in our own preliminary explorations.

For example, a list can be encoded by using {\tt :this} and {\tt :rest} keys on the same level of a V-value,
and having neurons with {\tt first}, {\tt rest}, and {\tt cons} activation functions. The only decision one
needs to make is whether to consider all lists as having infinite sequences of zeros at the tail, or whether
to incorporate an explicit list terminator (e.g. a keyword {\tt :end} mapped to {\tt 1}) in the formalism.

E.g. the following activation function for the neuron used to accumulate a list of ``interesting" events
is similar to the function used in our example of a DMM accumulating a list of mouse
clicks\footnote{\url{https://github.com/jsa-aerial/DMM/blob/master/examples/dmm/quil-controlled/jul_13_2017_experiment.clj}}:
\begin{verbatim}
(defn dmm-cons [accum-style-input]
  (let [old-self (get accum-style-input :self {})
        new-signal (get accum-style-input :signal {})]
    (if (interesting? new-signal)
      {:self {:this new-signal :rest old-self}}
      {:self old-self})))
\end{verbatim}

In order to maintain the accumulator metaphor, the {\tt :self} output and the {\tt :self} input of the
corresponding neuron are connected with weight 1.

Any linked structures which can be encoded inside the network matrix, can be encoded inside a similar
matrix not used as the network matrix (the only difference is that data structures encoded within the network
matrix tend to be ``active", as they are built over actively working neurons; this difference can potentially
be profound).

\subsection{From Programming via Composition of Transformers of Streams of V-values to Dataflow Matrix Machines}\label{sec:composition}

There is a rather long history of programming via composition of transformers of linear streams. In each of those cases,
linear streams can be represented as sufficiently general streams of V-values.

Perhaps the most well known example is the discipline of audio synthesis via composition of {\em unit generators},
which are transformers of streams of audio samples (streams of numbers, if one considers a single monophonic channel). That discipline
was created by Max Mathews in 1957 at Bell Labs~\cite{MVMathews}. It is typical for a modern audio synthesis
system to be crafted along those lines, even though the syntax can differ greatly. One of the classical textbooks
in that discipline is~\cite{PureDataRef}. We found the tutorial~\cite{Merz} on Beads, a realtime audio and music library for
Java and Processing, to be a convenient introductory text.

In this cycle of studies we explored a number of different examples of programming via composition of linear streams,
starting from dataflow programming of animations via composition of transformers of image streams~\cite{BMAlmostContinuous}.
During that series of experiments we discovered that if one does not want to impose the condition of the
dependency graph being acyclic, then one needs to maintain two elements of each stream at any given time,
the ``current" element, used by the transformers depending on that stream in their computations, and
the ``next" element, the element which is being computed. Then, after transformers computed their respective
``next" elements, there is a {\em shift} operation which makes all the ``next" elements current.

This cycle ``transform-shift" is a version of the two-stroke cycle used in this paper, and the shift operation is
what the ``down movement" of a DMM would look like, if all non-zero matrix rows would only have a single
non-zero element in each of them, and that element would be 1. In fact, any program built as a composition
of transformers of linear streams can be converted into an equivalent DMM by inserting a linear transformation
described by a matrix row with one non-zero element with weight 1 at each connection.

A number of examples of DMMs we explored in this cycle of studies, such as the character-processing DMM
described in Section 3 of~\cite{BMR3} or the DMM accumulating a list of mouse
clicks mentioned in Section~\ref{sec:datastruct} of the present paper, come from programs built as compositions of transformers of
linear streams converted to DMMs by inserting weight 1 connectors.

One aspect which used to be somewhat limited within the discipline of programming via composition of
transformers of linear streams was higher-order programming, i.e. transforming the transformers.
In particular, the higher-order constructions themselves were not expressed as transformations
of linear streams.

Dataflow matrix machines allow to continuously transform any
composition of stream transformers into any other composition of stream transformers,
and we present one way to do so {\em within the discipline of transforming linear streams} and
in a self-referential manner in the next section.

\section{Self-referential Mechanism}\label{sec:self-ref}

The ability to handle arbitrary linear streams implies the ability to handle streams of vectors shaped like
network connectivity matrices (be those flat two-dimensional matrices, or sparse multidimensional tensors
described in Section~\ref{sec:variadic}). This enables a rather straightforward mechanism to access and modify
the network matrix {\bf W}. We designate a neuron {\tt Self} emitting a stream of such matrices, and use the
most recent value from that stream as {\bf W} for the purpose of the next ``down movement" step.

We currently prefer to use an accumulator with $+$ activation function $y = x + \Delta x$ as {\tt Self} following~\cite{BMR4} (see Section~\ref{sec:lin-mult}).
{\tt Self} takes additive updates from other neurons in the network on its $\Delta x$ input, and other neurons
can take the stream of the current values of {\bf W} from the output of {\tt Self} making them aware of
the current state of the network connectivity.

Network self-modification based on the streams of network matrices was first introduced in~\cite{BMR2}, and
the principle of ``self-referential completeness of the DMM signature relative to the language
available to describe and edit the DMMs" was formulated there. That principle states that it is desirable
to have a sufficient variety of higher-order neurons to perform updates of the network matrix, so that any modifications
of a DMM (understood as a network or as a program) can be made by triggering an appropriate higher-order
neuron.

Paper~\cite{BMR3} explored ways for the network to modify itself by making deep copies of its own subgraphs.
The possibility of using self-referential matrix transformations as a new foundation for programming with linear streams,
somewhat similarly to lambda-calculus being a foundation for symbolic programming, was studied in~\cite{BMR4}.

Also, it was demonstrated in~\cite{BMR4} that this self-referential mechanism together with a few constant update
matrices gives rise to a wave pattern dynamically propagating in the network matrix; this result was also
verified in computer experiments (see Appendix B of~\cite{BukatinAnthony} for a more polished presentation).

However, all these studies were so far merely scratching the surface of what is possible with this mechanism.
In principle, it should allow the network to maintain an evolving population of its own subnetworks, to
maintain an evolving population of network update methods, to train network update methods as
a linear combination of available network update primitives, etc.

We hope that some of this potential will be explored in the future.

\section{Network Topology}\label{sec:network-topology}

The network topology, such as layers, is defined by the pattern of sparsity of {\bf W}: some of the connectivity
weights are kept at zero, and some are allowed to deviate from zero, and the network topology comes from that.

However, in the model of synchronous time we follow in this paper, all layers still work simultaneously. If it is desirable
for layers to work strictly in turn, one can use multiplicative mechanisms described in Section~\ref{sec:lin-mult} to orchestrate
the computations by turning
off the layers at the appropriate moments of time using zero {\em multiplicative masks}, and then by further optimizing
implementation to save processing time in such situations.

Not all weights which are non-zero need to be variable weights. It is often the case that
some of the non-zero weights are set to 1, and then it is the user's choice which of those
should be allowed to vary. 
The particularly frequent are the cases when the weight 1 in question is the only non-zero element of its matrix row.
The examples of cases where weights set to 1 occur naturally
include

\begin{itemize}
\item accumulators (Sections~\ref{sec:lin-mult},~\ref{sec:datastruct});
\item cases when a program expressed as composition of V-value transformers is translated into a DMM (Section~\ref{sec:composition});
\item special neural network topology\footnote{For example, let's express LSTMs and Gated Recurrent Unit networks
as networks built from sigmoid neurons, linear neurons, and neurons performing multiplication for gating
following Appendix C of~\cite{BMR4}, and let's assume that we are writing this in terms of neurons processing
scalar streams (streams of numbers). Then each of the two inputs for each neuron performing multiplication of
scalar streams is connected with weight 1 from a single output of an appropriate neuron.}.
\end{itemize}

\section{Subnetworks and Modularization}\label{sec:subnetworks}

The modern trend in artificial neural nets is to build the networks not from a large number of single neurons, but from
a relatively small number of modules such as layers, etc.

In this sense, it is convenient that neurons in DMMs are powerful enough to express the ``up movement" action of
whole subnetworks. This allows to build DMMs from a relatively small number of powerful neurons, if so desired.

In the old style DMMs~\cite{BMR1,BMR2,BMR3,BMR4}, the neurons had fixed arity and were powerful enough to express the ``up movement" action
of the subnetworks of fixed size. However, the networks themselves became variadic networks of unbounded size,
so the gap between single neurons and general subnetworks remained.

With variadic neurons this particular gap is eliminated.

In 2016, Andrey Radul formulated a principle stating that there is
no reason to distinguish between a neuron and a subnetwork, and
that it is a desirable property of a model to not have a difference
between a generalized neuron and a subnetwork.

The formalism of V-values and variadic neurons allows DMMs to
{\em fulfill this principle in the following limited sense}: single neurons are
powerful enough to express one ``up movement" action of any subnetwork
as one ``up movement" action of an appropriately crafted single neuron.

\section{Learning}\label{sec:learning}

There are various indications that dataflow matrix machines have strong potential for future machine learning applications.

DMMs contain well-known classes of neural networks with good machine learning properties, such as LSTM and Gated Recurrent Unit networks (Appendix C of~\cite{BMR4}).

At the same time, they allow to naturally express a number of various algorithms within a formalism which allows arbitrarily small modifications of programs,
where one can transform programs continuously by continuously transforming the matrices defining those programs.

The presence of well-developed self-referential facilities means that network training methods can be made part of the network itself,
making this a natural setting for a variety of ``learning to learn" scenarios.

DMM architecture is conductive to experiments with ``fast weights" (e.g.~\cite{BaHinton}).

Recently, we have been seeing very interesting suggestions that synthesis of small functional programs and
synthesis of neural network topology from a small number of modules might be closely related to each other~\cite{Olah, Nejati}.
The ability of single DMM neurons to represent neural network modules suggests that DMMs might provide the right
degree of generality to look at these classes of problems of network and program synthesis.

We are seeing evidence that syntactic shape of programs and their functionality provide sufficient information about
each other for that to be useful during program synthesis by machine learning methods (e.g.~\cite{SketchLearning}).
If a corpus of human-readable programs manually written in the DMM architecture emerges eventually, this
should be quite helpful for solving the problem of synthesis of human-readable programs performing
non-trivial tasks.

Given that DMMs form a very rich class of computational models, it makes sense to search for its various subclasses
for which more specialized methods of machine learning might be applicable.

\section{Historical Remarks and Related Work}\label{sec:Historical}

There are two ways one can arrive at the dataflow matrix machines. One can focus on programming with linear streams
and then notice that by interleaving linear and non-linear transformations of those streams, it is possible to obtain parametrization
of large classes of programs by matrices of numbers.

Another way is to focus on recurrent neural networks as a programming platform and to try to generalize them as much as
possible while retaining their key characteristic, which is parametrization by connectivity matrices of network weights.

In this section of the present paper we discuss some of the related work under both of these approaches.

\subsection{Recurrent Neural Networks as Programs}

It was recognized as early as 1940-s, that if one provides a neural network with a suitable model of unbounded memory
one obtains a Turing-universal formalism of computations~\cite{McCullochPitts}. Research studies formally establishing
Turing-universality for neural networks processing streams of reals include~\cite{Pollack, SiegelmannSontag}. 

More recent studies include such well-known approaches as~\cite{GravesWayneDanihelka, WestonChopraBordes}, which are
currently under active development.

Yet, as these approaches are gradually becoming more successful at learning neural approximations to known algorithms,
they do not seem to progress towards human-readable programs. In fact, it seems that while the expressive power of scalar-based neural networks
is sufficient to create Turing-complete esoteric programming languages, they are not expressive enough to become a pragmatic programming
platform. The restriction to scalar flows seems to either necessitate awkward encoding of complex data within reals (as in~\cite{SiegelmannSontag},
where binary expansions of real numbers are used as tapes of Turing machines), or to force people
to create networks depending in their size on data dimensionality and with any modularization and memory capabilities being external to the network
formalism, rather than being native to the networks in question.

The awkward encodings of complex data within reals hinder the ability to use self-modification schemas for scalar-based neural
networks. E.g. a remarkable early paper~\cite{Schmidhuber} has to use such encodings for addresses in the network matrix, and such encodings lead to very
high sensitivities to small changes of numbers involved, while the essence of correct neural-based computational schemas is
their robustness in the presence of noise.

Even such natural constructions within the scalar flow formalism as neurons with linear activation functions, such as identity, and neurons performing multiplication of
two arguments, each expressing a different linear combination of neuron outputs, encounter resistance in the field.

The power of linear
and multiplicative neurons was well understood at least as early as 1987~\cite{Pollack}. Yet, when the LSTMs were invented in
1997~\cite{LSTM}, the memory and gating mechanisms were understood as mechanisms external to neural networks, rather then the mechanisms
based straightforwardly on neurons with linear activation functions for memory and neurons performing multiplication for gating, which provide a natural way
to express memory and gating mechanisms in neural nets (see Appendix C of~\cite{BMR4}). 

Recently, the power of having linear activation functions, in particular identity, in the mix together with other activation functions
is finally getting some of the recognition it deserves (the paper~\cite{KaimingHe} is now a well-cited paper). However, the explicit activation functions
of arity two, such as multiplication, are still quite exotic and often difficult to explicitly incorporate into existing software frameworks for neural networks.

We think that dataflow matrix machines as presented here, with their vector flows and multiple arities for activation functions,
 provide the natural degree of generality for neural networks.

\subsection{Programming with Linear Streams}

Continuous computations (which tend to be computations with linear streams) have a long history, starting with electronic
analog computers. The programs, however, were quite discrete: a pair of single-contact sockets was either connected with
a patch cord, or it was not connected with a patch cord.

More modern dataflow architectures focusing on work with linear streams representing continuous data include, for example,
LabVIEW~\cite{LabViewRef} and Pure Data (e.g.~\cite{PureDataRef}). The programs themselves are still quite discrete.

To incorporate higher-order programming methods within the paradigm of programming with linear streams, 
the space of programs themselves needs to become continuous.
Neural networks represent a step in this direction. While the network topology itself is discrete (the connection between
nodes is either present, or not), when one expresses the network topology via its weight-connectivity matrix, the
degree to which any particular edge is present (the absolute value of the weight associated with it) can be made
as small as desired, and this provides the continuity we are after.

The particular line of work we are presenting in this paper emerged in 2012-2013, when it was recognized by our group that the
approximation domains providing continuous denotational semantics in the theory of programming languages
can acquire the structure of vectors spaces, when equipped with cancellation properties missing in the standard theory
of interval numbers.
Namely, there must be enough overdefined (partially inconsistent) elements
in those spaces to produce zero on addition by the mechanism of cancellation with underdefined (partially defined)
elements.
For interval numbers, this corresponds to introduction of pseudosegments $[a, b]$ with the contradictory property that $b<a$, following
Warmus~\cite{Warmus}.
For probabilistic spaces, this corresponds to allowing negative values of probabilities in addition to usual non-negative values.
The mathematics of the resulting {\em partial inconsistency landscape} is presented in Section 4 of~\cite{BM2015}.

By 2015 it became apparent to our group that programming with linear streams was a rich formalism which included programming
with streams of probabilistic samples and programming with {\em generalized animations}. This framework seemed to provide
methods for {\em continuous higher-order programming}, and, moreover, it had good potential for obtaining more
efficient schemas for genetic programming by allowing to introduce the motives similar to {\em regulation of gene expression}
into genetic programming frameworks~\cite{BM2015}. Crucially, it also became apparent at that time that if one introduced the discipline
of interleaving linear and non-linear transformations of linear streams, then one could parametrize large classes of programs by
matrices of numbers~\cite{BukatinMatthewsDataflowGraphsAsMatrices, BM2015}.

The first open-source software prototypes associated with this approach also appeared in 2015~\cite{Fluid}.

In 2016 we understood that the resulting formalism generalized recurrent neural networks, and the term {\em dataflow matrix machines}
was coined~\cite{BMR1}. The modern version of the self-referential mechanism in DMMs and the first precise description of how
dataflow matrix machines function, given that their matrices can be dynamically expanded, appeared in~\cite{BMR2}.
The programming patterns for the resulting software framework were studied in~\cite{BMR3}. A more theoretical paper~\cite{BMR4} explored
the possibility of using self-referential matrix transformations instead of lambda-calculus as the foundation in this context and established further
connections between neural networks and the mathematics of the partial inconsistency landscape.  

The formalism of finite prefix trees with numerical leaves (the vector space of recurrent maps) was introduced in the Fall of 2016.
This formalism was inspired by our work with Clojure programming language~\cite{Hickey, Clojure}.
The first open-source implementation of a version of DMMs based on that formalism and written in Clojure was produced
in that time frame~\cite{DMM}, and a research paper based on this architecture was presented recently at LearnAut 2017~\cite{BukatinAnthony}.

\section*{Acknowledgments}

We would like to thank Dima-David Datjko, Elena Machkasova, and Elena Nekludova for helpful discussions of the material presented in~\cite{BukatinAnthony} and here.


\begin{thebibliography}{1}

\bibitem{BaHinton}J.~Ba, G.~Hinton, V.~Mnih, J.~Leibo, C.~Ionescu. {\em Using Fast Weights to Attend to the Recent Past}, October 2016.\\
\url{https://arxiv.org/abs/1610.06258}

\bibitem{BukatinAnthony} M.~Bukatin, J.~Anthony. {\em Dataflow Matrix Machines as a Model of Computations with Linear Streams}. In LearnAut 2017 (``Learning and Automata" Workshop at LICS 2017). \url{https://arxiv.org/abs/1706.00648}

\bibitem{BM2015}
M.~Bukatin, S.~Matthews. {\em Linear Models of Computation and
Program Learning.} In G. Gottlob et al., editors, GCAI 2015, EasyChair
Proceedings in Computing, vol. 36, pages 66--78, 2015.\\
\url{https://easychair.org/publications/paper/Q4lW}

\bibitem{BMAlmostContinuous}
 M.~Bukatin, S.~Matthews. {\em Almost Continuous Transformations of Software and Higher-Order
Dataflow Programming}. Preprint, July 2015.\\ \url{http://arxiv.org/abs/1601.00713}

\bibitem{BukatinMatthewsDataflowGraphsAsMatrices} M.~Bukatin, S.~Matthews. {\em Dataflow Graphs as Matrices and Programming with Higher-order Matrix Elements}.
Preprint, August 2015.\\ \url{https://arxiv.org/abs/1601.01050}

\bibitem{BMR1}
M.~Bukatin, S.~Matthews, A.~Radul. {\em Dataflow Matrix Machines as a Generalization of Recurrent
Neural Networks}, March 2016.\\
\url{http://arxiv.org/abs/1603.09002}

\bibitem{BMR2}
M.~Bukatin, S.~Matthews, A.~Radul. {\em Dataflow Matrix Machines as
Programmable,  Dynamically  Expandable,  Self-referential
Generalized Recurrent Neural Networks}, May 2016.
\url{https://arxiv.org/abs/1605.05296}

\bibitem{BMR3}
M.~Bukatin, S.~Matthews, A.~Radul. {\em  Programming  Patterns  in
Dataflow Matrix Machines and Generalized Recurrent Neural Nets}, June 2016.\\
\url{https://arxiv.org/abs/1606.09470}

\bibitem{BMR4}
M.~Bukatin, S.~Matthews, A.~Radul. {\em  Notes on Pure Dataflow Matrix Machines: Programming with 
Self-referential Matrix Transformations}, October 2016.
\url{https://arxiv.org/abs/1610.00831}

\bibitem{Clojure}
Clojure programming language. \url{https://clojure.org}

\bibitem{DMM}
DMM project GitHub repository, 2016-2017.\\
\url{https://github.com/jsa-aerial/DMM}

\bibitem{PureDataRef}
A. Farnell. {\em Designing Sound}. MIT Press, 2010.

\bibitem{Fluid}
Fluid: Project ``Fluid" GitHub repository, 2015-2017.\\
\url{https://github.com/anhinga/fluid}

\bibitem{GravesWayneDanihelka}
A.~Graves, G.~Wayne, I.~Danihelka. {\em Neural Turing Machines},
October 2014.
\url{https://arxiv.org/abs/1410.5401}

\bibitem{Hickey}
R.~Hickey. {\em The Clojure Programming Language}. In Proceedings of the 2008 Symposium on Dynamic Languages,
Association for Computing Machinery.

\bibitem{LabViewRef}
W.~Johnston, J.~Hanna, R.~Millar. Advances in Dataflow Programming
Languages. {\em ACM Computing Surveys}, 36:1--34, 2004.

\bibitem{KaimingHe}
K.~He, X.~Zhang, S.~Ren, J.~Sun. {\em Identity Mappings in Deep Residual Networks}, March 2016.
\url{https://arxiv.org/abs/1603.05027}

\bibitem{LSTM}
S.~Hochreiter, J.~Schmidhuber. Long Short-Term Memory. {\em Neural Computation}, 9(8):1735--1780, 1997. 

\bibitem{Karpathy}
A.~Karpathy. {\em The Unreasonable Effectiveness of Recurrent Neural Networks}, May 2015.\\
\url{http://karpathy.github.io/2015/05/21/rnn-effectiveness/}

\bibitem{Kornai} A.~Kornai. {\em Mathematical Linguistics}. Springer-Verlag, 2008.

\bibitem{MVMathews}
M.~Mathews. The Digital Computer as a Musical Instrument. {\em Science}, 142: 553--557, 1963. 

\bibitem{McCullochPitts}
W.~McCulloch, W.~Pitts. A Logical Calculus of the Ideas Immanent in Nervous Activity. {\em The Bulletin of Mathematical Biophysics}, 5:115--133, 1943.

\bibitem{Merz}
E.~Merz. {\em Sonifying Processing: The Beads Tutorial}. 2011.\\
\url{http://evanxmerz.com/?page_id=18}

\bibitem{Mikolov}
T.~Mikolov, I.~Sutskever, K.~Chen, G.~Corrado, J.~Dean. {\em Distributed Representations of Words and Phrases and their Compositionality},
October 2013. \url{https://arxiv.org/abs/1310.4546}

\bibitem{SketchLearning}
V.~Murali, S.~Chaudhuri, C.~Jermaine. {\em Bayesian Sketch Learning for Program Synthesis}, March 2017.
\url{https://arxiv.org/abs/1703.05698}


\bibitem{Nejati}
A.~Nejati. {\em Differentiable Programming}, August 2016.\\
{\footnotesize \url{https://pseudoprofound.wordpress.com/2016/08/03/differentiable-programming}}

\bibitem{Olah}
C.~Olah. {\em Neural Networks, Types, and Functional Programming}, September 2015.
\url{http://colah.github.io/posts/2015-09-NN-Types-FP}

\bibitem{Pollack} J.~Pollack.
{\em On Connectionist Models of Natural Language Processing.}
PhD thesis, University of Illinois at Urbana-Champaign, 1987.
Chapter 4 is available at
\url{http://www.demo.cs.brandeis.edu/papers/neuring.pdf}

\bibitem{Schmidhuber}
J.~Schmidhuber. {\em A ``Self-Referential" Weight Matrix}, In: S.~Gielen, B.~Kappen, eds., ICANN ’93: Proceedings
of the International Conference on Artificial Neural Networks, Springer, 1993, pp. 446--450.

\bibitem{SiegelmannSontag}
H.~Siegelmann, E.~Sontag. On the Computational Power
of Neural Nets.
{\em Journal of Computer and System Sciences}, 50:132--150, 1995.

\bibitem{Warmus}
M.~Warmus. Calculus of Approximations. {\em Bull. Acad. Pol. Sci., Cl. III}, {\bf 4}(5):253--259, 1956.\\
\url{http://www.cs.utep.edu/interval-comp/warmus.pdf}

\bibitem{WestonChopraBordes}
J.~Weston, S.~Chopra, A.~Bordes. {\em Memory Networks}, October 2014.\\
\url{https://arxiv.org/abs/1410.3916}

\end{thebibliography}
\end{document}